\documentclass{article}

\usepackage{arxiv}

\usepackage[utf8]{inputenc} 
\usepackage[T1]{fontenc}    
\usepackage{hyperref}       
\usepackage{url}            
\usepackage{booktabs}       
\usepackage{amsfonts}       
\usepackage{nicefrac}       
\usepackage{microtype}      
\usepackage{lipsum}		
\usepackage{graphicx}
\usepackage{doi}
\usepackage{caption}
\usepackage{subcaption}
\usepackage{cite}
\usepackage{amsmath,amsfonts}
\usepackage{mathtools}

\DeclarePairedDelimiter\floor{\lfloor}{\rfloor}
\usepackage{algorithmic}
\usepackage{graphicx}
\usepackage{textcomp}
\usepackage{xcolor}
\usepackage{comment}
\def\BibTeX{{\rm B\kern-.05em{\sc i\kern-.025em b}\kern-.08em
    T\kern-.1667em\lower.7ex\hbox{E}\kern-.125emX}}

\title{A Study on Tiny YOLO for Resource Constrained XRay Threat Detection}

\author{ \href{https://orcid.org/0000-0000-0000-0000}{\includegraphics[scale=0.06]{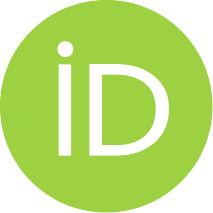}\hspace{1mm}Venkata Raghav Ambati}\thanks{Use footnote for providing further
		information about author (webpage, alternative
		address)---\emph{not} for acknowledging funding agencies.} \\
	Department of Mathematics\\
	IIT Hyderabad\\
	\texttt{venkata.raghav07@gmail.com} \\
	\And
	\href{https://orcid.org/0000-0000-0000-0000}{\includegraphics[scale=0.06]{orcid.pdf}\hspace{1mm}Ayon Borthakur} \\
	Department of Artificial Intelligence\\
	IIT Hyderabad\\
	\texttt{ayon.borthakur@ai.iith.ac.in} \\
}

\renewcommand{\shorttitle}{\textit{arXiv} Template}

\hypersetup{
pdftitle={A template for the arxiv style},
pdfsubject={q-bio.NC, q-bio.QM},
pdfauthor={Venkata Raghav Ambati, Ayon Borthakur},
pdfkeywords={First keyword, Second keyword, More},
}

\begin{document}
\maketitle

\begin{abstract}
\end{abstract}

This paper implements and analyses multiple nets to determine their suitability for edge devices to solve the problem of detecting Threat Objects from X-ray security imaging data. There has been ongoing research on applying Deep Learning techniques to solve this problem automatedly. We utilize an alternative activation function calculated to have zero expected conversion error with the activation of a spiking activation function, in the our tiny YOLOv7 model. This \textit{QCFS} version of the tiny YOLO replicates the activation of ultra-low latency and high-efficiency SNN architecture and achieves state-of-the-art performance on CLCXray which is another open-source XRay Threat Detection dataset, hence making improvements in the field of using spiking for object detection. We also analyze the performance of a Spiking YOLO network by converting our QCFS network into a Spiking Network.


\section{Introduction}

X-ray threat Detection is necessary for maintaining public safety in airports, subways, and train stations. For the longest time, the approach to this problem involved manual screening by experienced personnel of the security staff. However, this approach is prone to human error and requires some labor. There have been many attempts to automate this task using Deep Learning models by using various pre-trained state-of-the-art algorithms such as RCNNs and YOLO [1]. This threat detection task becomes much more complicated under resource-constrained environments due to memory, battery power, and latency requirements such as in a handheld X-ray device. 

However recently, with increasing interest in the field of Spiking Neural Networks trying to be used for solving Object Detection problems, there have been many proposed spiking-based object detection models. Object detection itself is a very fundamental task in Computer Vision and ANNs currently dominate the market for the algorithms to be used for this task. However, they have high energy consumption making them difficult to deploy on mobile devices. SNNs are third-generation neural networks, supposed to replicate the action of how our brain receives and transfers messages using the neural pathways in our nervous system. SNNs transmit information through spike sequences, which are sparse in nature. Due to this sparsity, SNNs have remarkable energy efficiency and are the go-to choice for Neuromorphic Chips. There are still quite a few developments to be made till SNNs are used as regularly as ANNs for tasks like Object Detection. However, this is potentially one of them.
One issue with SNNs arises due to the non-differentiability of the equations governing the network, causing an issue with loss calculation. One way to get past this is to convert pre-trained ANN weights into an SNN network. The other is to train SNN weights from scratch using STDP or STDB [6]. The work in this paper features a technique from the former. There have been some efforts to use an SNN network for object detection. [5] implements a Spiking YOLO network (on YOLOv3, a widely used object detection algorithm back then) and achieves remarkable results that are comparable (up to 98\%) to Tiny YOLO (a more compact and efficient version of YOLO).YOLOv7 [2] is a state-of-the-art ANN algorithm, pre-trained on the MS COCO dataset and finetuned for object detection. Inspired by the works of [5] and [3], we implement the Quantization Clip-Floor-Shift Activation Function, which approximates (with zero expectation error) the activation of a Neuron for Tiny YOLOv7. YOLO has never been used in XRay Threat Detection, so we also achieved remarkable results in the Transfer Learning problem (considering mAP and F1 scores). We also analyze the problem of converting the ANN YOLO network into a Spiking Neural Network by converting a pre-trained QCFS network into an SNN network.

\section{Related Work and Methods}

\subsection{Dataset and Metrics}
CLCXray [4] is an open source XRay security baggage dataset consisting of multiple Threat and Non-Threat items found in the real subway security baggage screening process. Consisting of 12 classes (Fig 1). We train Tiny YOLOv7 (and its QCFS counterpart) on 7,652 images, conducting test and validation with 955 images each (an 80-10-10 split) with images consisting of baggage with items shown in (Fig 1). Precision, Recall, Mean Average Precision, and F1 scores are used as metrics. The data was pre-processed and converted into YOLO format. 

\begin{figure}[!ht]
\centering
    \includegraphics[scale=.25]{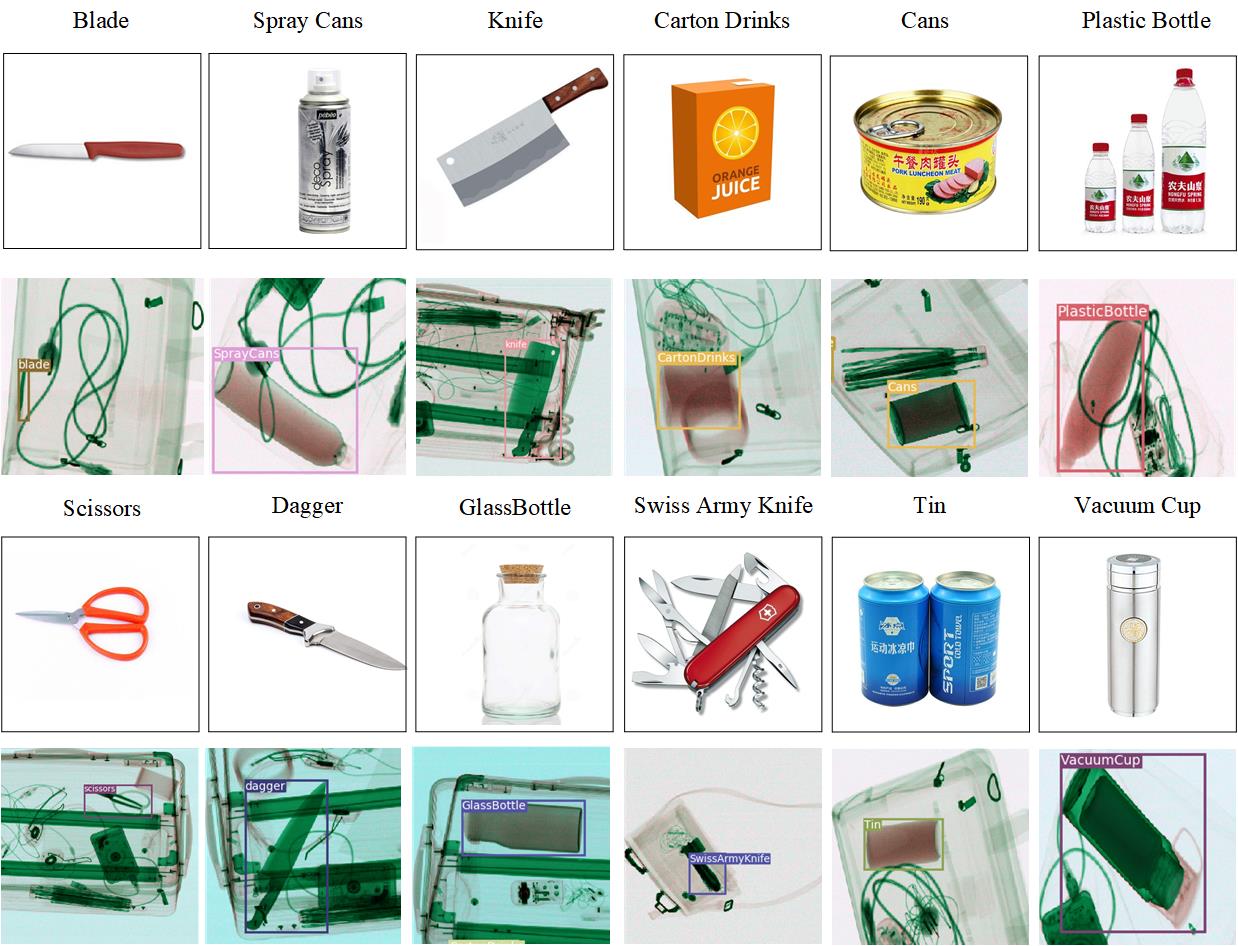}
    \caption{The CLCXray dataset has 5 types of cutters and 7 liquid containers}
    \label{fig:data}
\centering
\end{figure}

\subsection{Tiny YOLO model}
YOLOv7 [2] is a real-time object detector using the Extended Efficient Layer Aggregation Network (E-ELAN) architecture - based on the ELAN architecture. It also has a Trainable Bag-of-Freebies for better training cost and model accuracy. Tiny YOLOv7 is a smaller YOLOv7 model optimized for edge devices. While YOLOv7 has 106 Conv, Sequential, Concat, Detect, and MaxPooling layers using Sigmoid Linear Unit (SiLU) as the activation function, Tiny YOLOv7 has 78 layers using Leaky Rectified Linear Unit (Leaky ReLU) as the activation function. YOLOv7 improves upon earlier models in terms of accuracy but has increased training costs. The Backbone, Neck, and Head of the model are shown in the figure (Fig 2). The BConv layers have a 2D Convolutional Layer, an Activation Layer with ReLU (SiLU for YOLOv7) as the function, and a Batch Normalisation Layer (Fig 3). The MPConv layer has both BConv and MaxPooling Layers. We use Quantization Clip-Floor-Shift Activation in the Activation Layer instead of ReLU and replace the MaxPool layers with AvgPool layers [3].

\begin{figure}[!ht]
\centering
    \includegraphics[scale=.25]{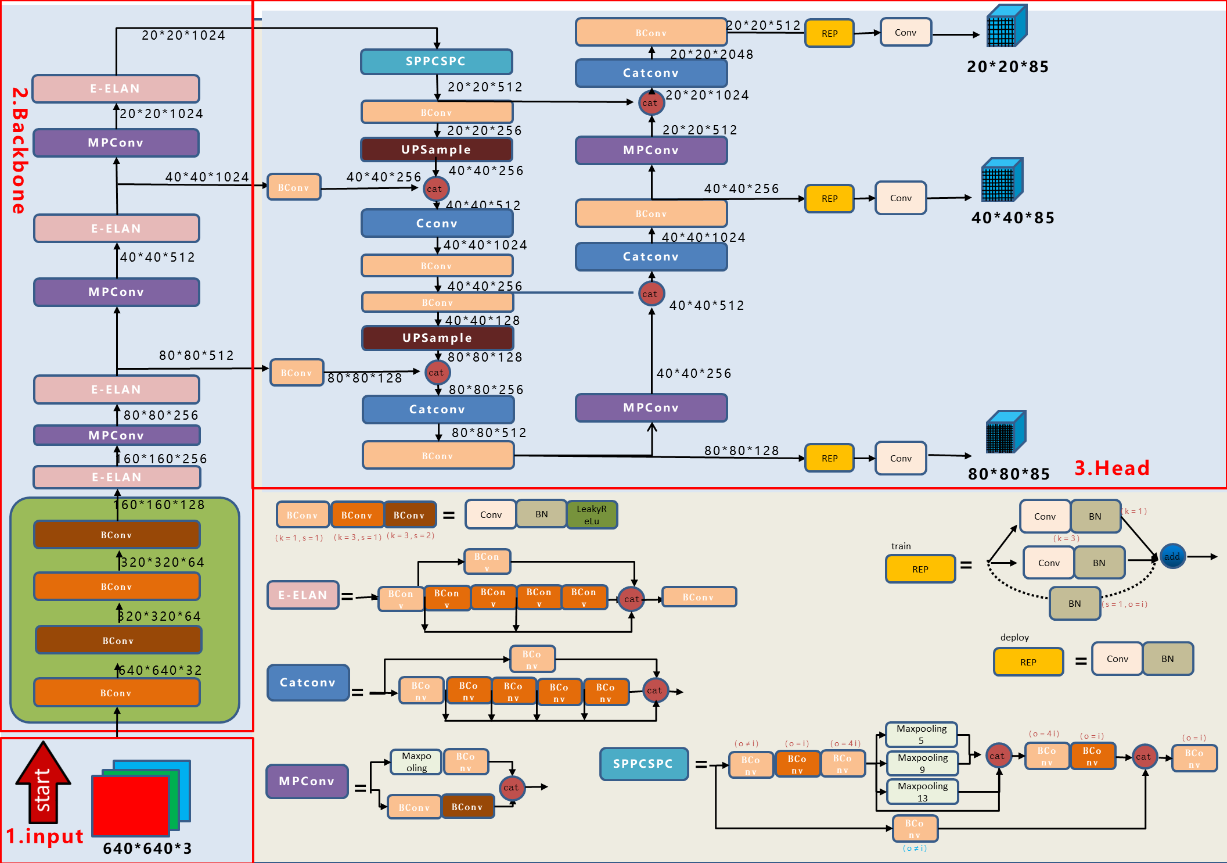}
    \caption{Overview of YOLOv7 Architecture. It uses E-ELAN architecture. The input is taken as dimensions of the input image. The E-ELAN architecture is the computational part of the backbone of YOLOv7.}
    \label{fig:model}
\centering
\end{figure}

\subsection{Transfer Learning}
The problem of automating the process of screening baggage in security using XRay scans has seen a lot of research [1][7] with the usage of RCNN and YOLO series being the most common of CNN-based detectors. YOLO is different from RCNN in that YOLO does not select targets by generating candidate frames but performs target prediction and recognition directly at the pixel level. Images of scanned security baggage are given and YOLO draws a bounding box around the objects from a class and assigns it a confidence value (Fig 4). The (Tiny) YOLOv7 model is pre-trained using the MS COCO dataset. These weights are then trained on the CLCXray dataset using the aforementioned data split for the problem of X-ray threat Detection. Both the ANN and QCFS models perform quite well on the task of detecting objects in X-ray scans (as shown in Section III).
\begin{figure}
\centering
    \includegraphics[width=.45\textwidth]{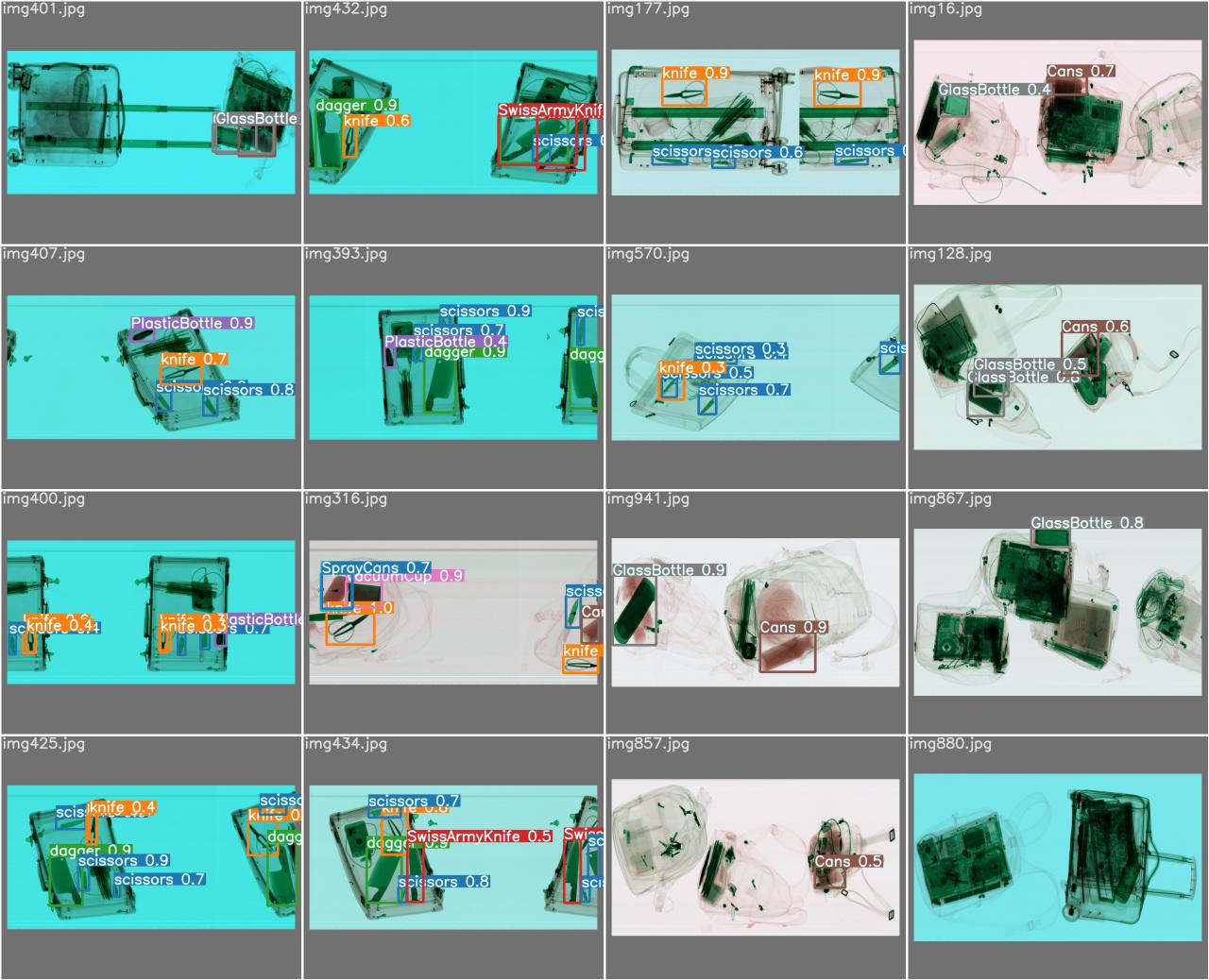}
    \caption{Input images that are fed to the algorithm. The model assigns a confidence value after creating a bounding box around the object.}
    \label{fig:labelled images}
\centering
\end{figure}
 
\subsection{QCFS and SNN Conversion}
Tiny YOLOv7 uses ReLU as its activation function. [3] notes that on replacing ReLU with QCFS for activation, the conversion error for ANN-SNN can be approximated as zero for any arbitrary value of timestep T. The conversion error for the $l^{th}$ layer (denoted by $\hat{e_l}$ for the $l^{th}$ layer) between ANN and SNN is given by - 
\begin{equation}
    \hat{e_l} \:=\: \phi^l(T) - \:a^l
\end{equation}
\begin{equation}
        = \; \theta^l\:clip(\frac{1}{T}\floor*{\frac{z^lT + v^l(0)}{\theta^l}}) \\ - \; \lambda^l\:clip(\frac{1}{L}\floor*{\frac{z^lL}{\lambda^l}}, 0, 1)
\end{equation}
\begin{equation}
    = 0
\end{equation}
Where $\varphi^l(T)$ is the output of the SNNs in the range of [0, $\theta^l$], given by -
\begin{equation}
    \varphi^l(T) = \frac{\Sigma_{i = 1}^{T}\:x^l(i)}{T}
\end{equation}
Where $x^l$ is written as a function $\theta^l$ (the unweighted synaptic potential) as -
\begin{equation}
    x^l(t) = s^l(t)\theta^l
\end{equation}

And $a^l$ is the output of the ANN with h as the activation function. In our case, we define our activation function 'h' - the Quantization Clip-Floor-Shift Activation function as follows -
\begin{equation}
    a^l = h(W^la^{l - 1}), \;\;\; l = 1, 2..., M
\end{equation}
or
\begin{equation}
    a^l = h\hat(z^l) 
\end{equation}
\begin{equation}
    = \: \lambda^lclip(\frac{1}{L}\floor*{\frac{z^lL}{\lambda^l} + \varphi},\: 0,\: 1)
\end{equation}

It turns out that the Expectation of this conversion error cancels out to be 0 for any arbitrary total timestep 'T' and the equivalent 'Quantization Step' L (a hyperparameter) for the ANN, if our hyperparameter $\varphi$ - that controls the shift of the activation function in the source ANN is equal to $\frac{1}{2}$, i.e
\begin{equation}
    \forall\:T, L \;\;\;\;\;\;\; E_z(\hat{e_l})|_{\varphi=\frac{1}{2}} = 0.
\end{equation}
Therefore, L is the only undetermined hyperparameter in this method. This result allows us to achieve high performance at ultra-low timesteps. 'L' can be changed based on the accuracy vs performance trade-off. Essentially, we replace the ReLU function in our BConv layer with QCFS (Fig 5) and also replace the Max Pooling layers with Average Pooling (*). Training with QCFS is not as straightforward as ReLU ANN. We use the straight-through estimator(*) with the following derivation rule -
\begin{equation}
\frac{\partial\hat{h_i}(z^l)}{\partial{z^l_i}} \: = 
    \begin{cases}
         1 & \text {, } \frac{-\lambda^l}{2L} < z_i^{l} < \lambda^l - \frac{-\lambda^l}{2L}\\
         0 & \text {, otherwise }
    \end{cases}
\end{equation}
\begin{equation}
\frac{\partial\hat{h_i}(z^l)}{\partial{\lambda^l_i}} \: =
    \begin{cases}
        \frac{\hat{h_i}(z^l) - z_i^{l}}{\lambda^l} & \text {, } \frac{-\lambda^l}{2L} < z_i^{l} < \lambda^l - \frac{-\lambda^l}{2L}\\
        0 & \text {, } z_i^{l} < \frac{-\lambda^l}{2L}\\
        1 & \text {, } z_i^{l} \geq \lambda^l - \frac{-\lambda^l}{2L}
    \end{cases}
\end{equation}

Then we perform Stochastic Gradient Descent to train the overall algorithm.

We train the model after converting ReLU into QCFS and then replace QCFS in the trained weights with a Neuron for ANN-SNN conversion. We compare two different conversions, the first model with just the first activation being replaced by a neuron and the other one with only the last activation being replaced by a neuron.

\begin{figure}
    \centering
    \includegraphics[scale=.35]{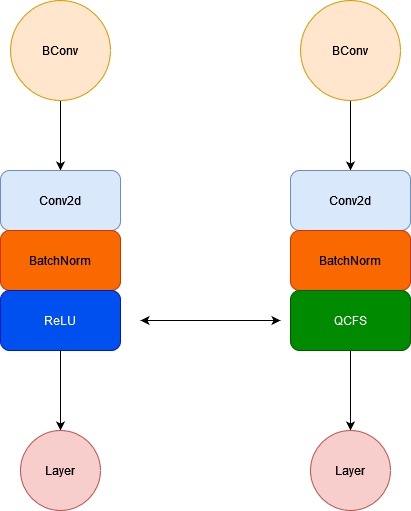}
    \caption{The BConv Layer of YOLOv7 consists of a conv2d layer, a batch norm layer, and an activation layer. For our Tiny YOLO QCFS Model, we replace the ReLU activation with QCFS.}
    \label{fig:bconv}
\centering
\end{figure}

\section{Emperical Results}

\label{sec:others}

\subsection{Evaluating Metrics}

Training tiny YOLOv7 on the CLCXray dataset yielded a mAP@.5 of around .808 and a mAP@.5:.95 of .622 with the validation data. When we replaced its ReLU layers with QCFS and MaxPool layers with AvgPool (according to the pseudocode in [3], with a different loss function) we achieved a mAP@.5 of .836 and mAP@.5:.95 of .648 at 4 Quantization Steps. P-R Curves are shown below (Fig 5). Comparing the F1 scores of both models, the ANN model has an F1 score of .80 at a Confidence Threshold of .486, while the QCFS model has an F1 score of .82 at a Confidence Threshold of .576. Both the F1 curves are shown below (Fig 6). The P and R curves of both models are also plotted below (Fig 4 and 5). Comparing the loss curves for both models, the loss at convergence is slightly less for the ANN model (0.041 vs .076). The loss curves are plotted below (Fig 6)

\begin{figure}
     \centering
     \begin{subfigure}[b]{0.35\textwidth}
         \centering
         \includegraphics[width=\textwidth]{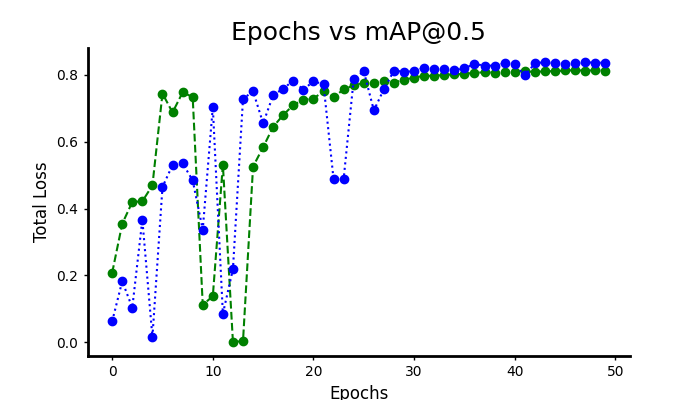}
         \caption{Convergence of mAP@.5. QCFS converges to a value slightly more than the ANN model (.836 vs .813)}
         \label{fig:Epochs vs mAP@.5}
     \end{subfigure}
     \hfill
     \begin{subfigure}[b]{0.35\textwidth}
         \centering
         \includegraphics[width=\textwidth]{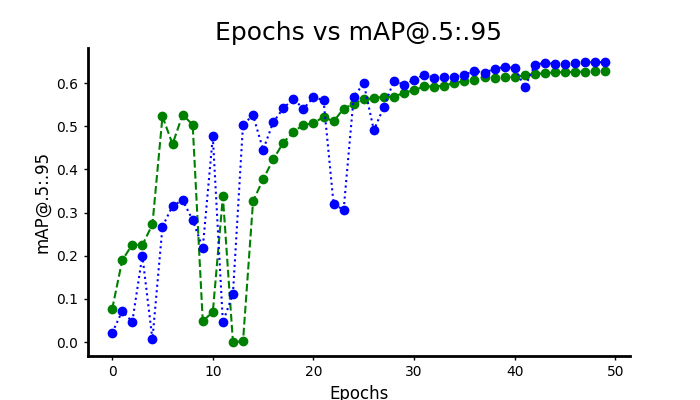}
         \caption{Convergence of mAP@.5:.95. QCFS converges to a value slightly more than the ANN model (.649 vs .628)}
         \label{fig:Epochs vs mAP@.5:.95}
     \end{subfigure}
     \hfill
      \begin{subfigure}[b]{0.35\textwidth}
         \centering
         \includegraphics[width=\textwidth]{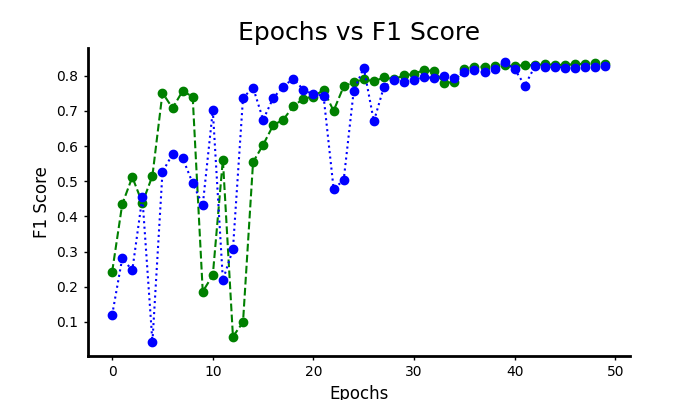}
         \caption{Convergence of F1 score. QCFS converges to a value slightly more than the ANN model (.82 vs .80)}
         \label{fig:Epochs vs F1 score}
     \end{subfigure}
      \begin{subfigure}[b]{0.35\textwidth}
         \centering
         \includegraphics[width=\textwidth]{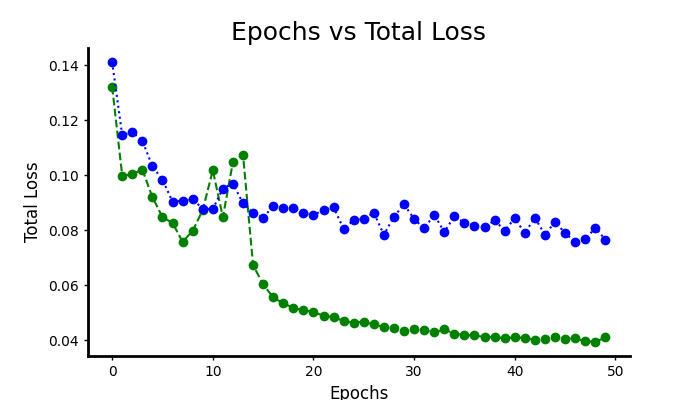}
         \caption{Total Loss (Box Loss + Objectness Loss + Classification Loss) vs Number of Epochs. ANN model converges a smoother and converges ato a slightly lower value (0.076 vs 0.041).}
         \label{fig:Epochs vs Total Loss}
     \end{subfigure}
        \caption{The Blue dotted line is QCFS and the Green Dashed line is the ANN model. Training for 50 epochs. We can conclude that there is marginal, albeit noticeable improvement with QCFS (optimized for edge devices) in all the performance metrics, while having only a slightly higher total loss.}
        \label{fig:Metrics}
\end{figure}

\subsection{Test results}
The ANN Tiny YOLO model has a mAP@.5 of .616 and is slightly outperformed by our QCFS model which has a score of .633. Comparing F1 scores, the models perform very similarly with the ANN model having a score of .60 at a Confidence Threshold of .671 while the QCFS model has a slightly higher score of .62 at a slightly lower Confidence Threshold of 0.524. The P and R curves of the test results of both models are also plotted below (Fig 6). 

The QCFS activation function performs with direct replacement of the YOLOv7 weights, with no other re-parameterization. \textit{There is also a possibility of getting binary outputs equivalent to actual spikes and using the quantization step to replicate an SNN even more closely}.

\subsection{SNN Conversion}

The mAP@.5 of the SNN model in which the QCFS' first activation is converted to a Neuron is <0.01. The mAP of the other model with the last QCFS activation converted to a Neuron performs reasonably better with a mAP@.5 of .375 for all classes, although the drop in performance when the neuron replaces the QCFS activation at the end is noticeable. The last layer neuron model has its mAP graph shown below (Fig 7). One possible reason could be the fact that the SNN activation outputs binary values, rather than the floating point expected by the YOLO architecture. This also could lead to a potential loss of information in the final output, hence causing the network to have lesser confidence.

\begin{figure}
     \centering
     \begin{subfigure}[b]{0.235\textwidth}
         \centering
         \includegraphics[width=\textwidth]{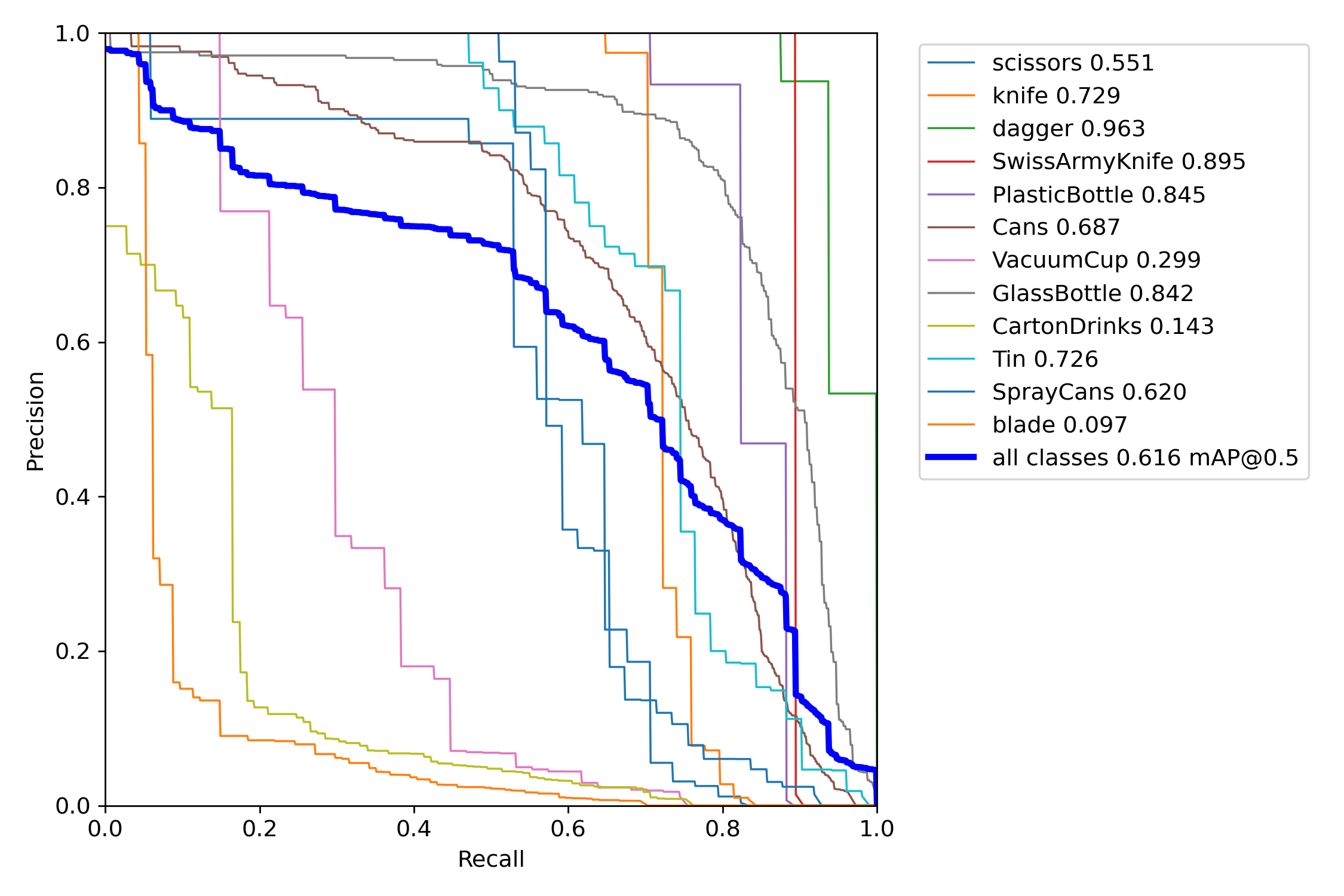}
         \caption{P-R Curve for ANN}
         \label{fig:PR Curve ANN Test}
     \end{subfigure}
     \hfill
     \begin{subfigure}[b]{0.235\textwidth}
         \centering
         \includegraphics[width=\textwidth]{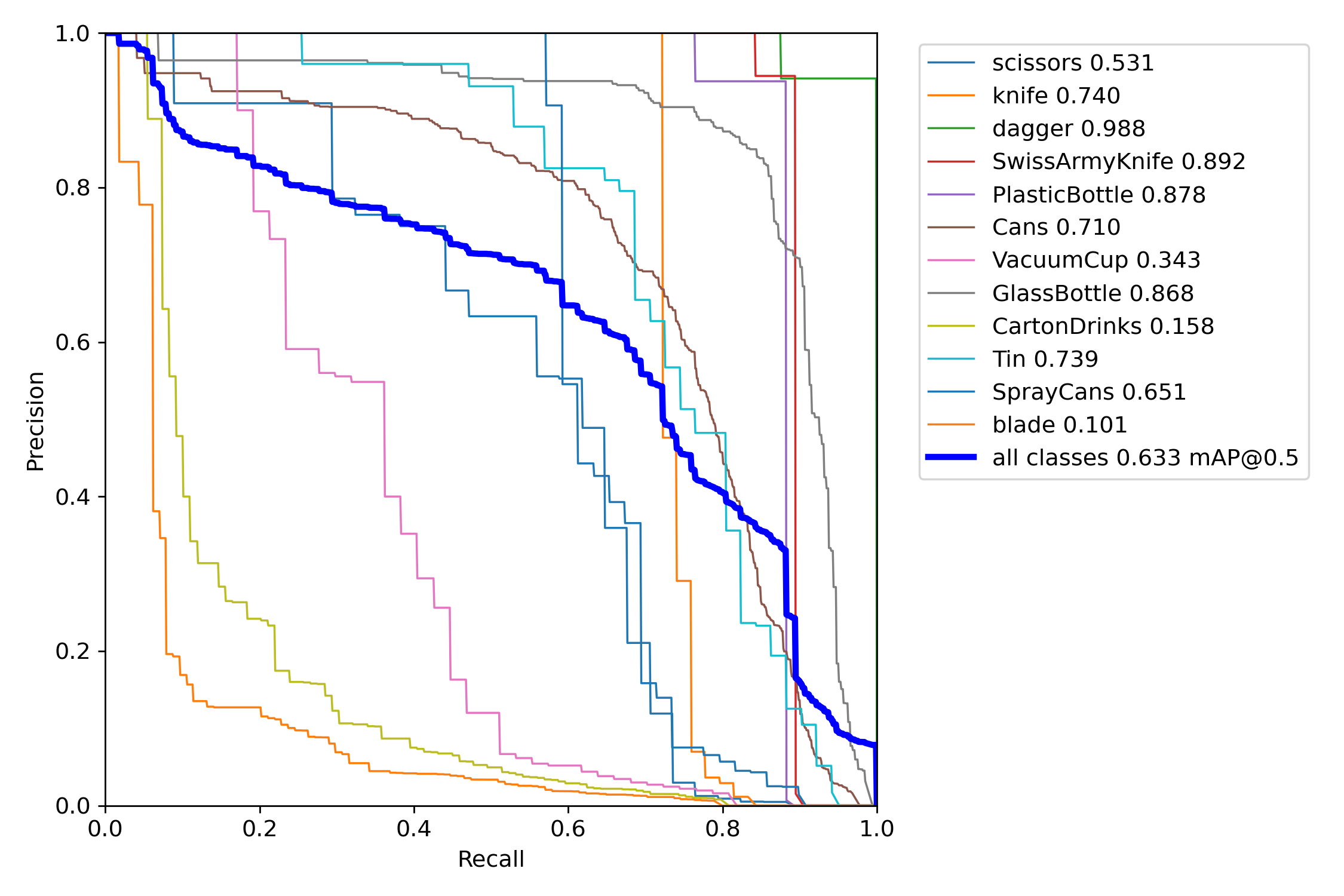}
         \caption{P-R Curve for QCFS}
         \label{fig:PR Curve QCFS Test}
     \end{subfigure}
     \hfill
        \caption{Comparing PR Curve of the test set of both the models. The QCFS model has slightly better mAP@.5 (.633 vs .616) than the ANN model.}
        \label{fig:PR Curves}
\end{figure}


\begin{figure}
\centering
    \includegraphics[width=.25\textwidth]{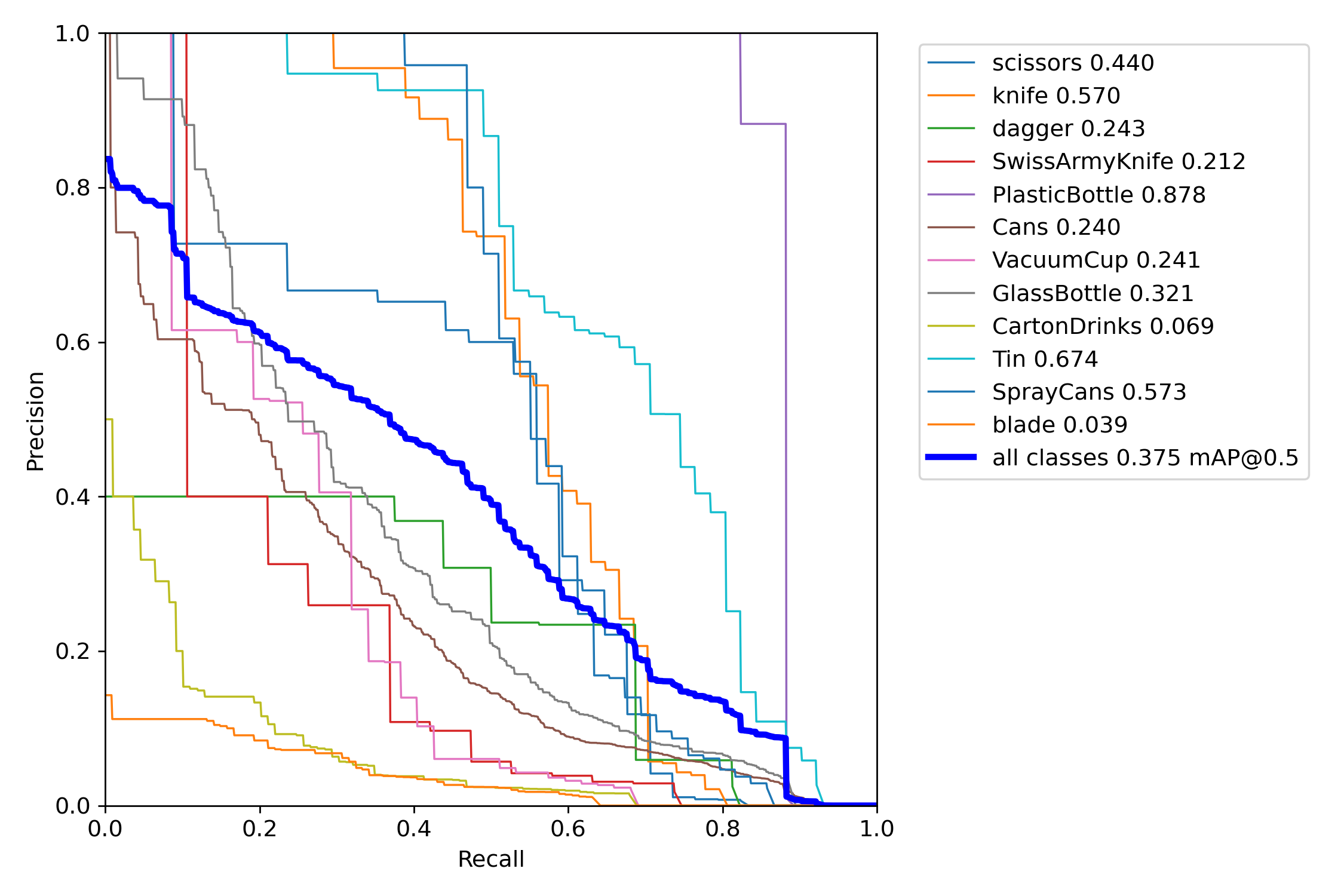}
    \caption{P-R curve for the model with its last QCFS activation replaced by Neuron.}
    \label{fig:PR Curve SNN Last Layer}
\centering
\end{figure}

\section{Conclusion}

We have trained a Tiny YOLOv7 model on an Open-Source XRay Baggage Threat dataset. We have also implemented a YOLOv7 model with Quantization Clip-Floor-Shift Function which is used as the activation function and has zero expected conversion error with an equivalent SNN network's activation. This allows us to approximate an ultra-low latency SNN (around 4-16 time-steps) version of the standard Tiny YOLOv7 model, which is optimized for edge computing devices. The obtained QCFS Tiny YOLOv7 model performs better than the standard Tiny YOLOv7 by some margin. We also analyze the problem of converting pre-trained ANN weights to an SNN. All the model metrics can be seen in the table below (Table 1).

\begin{center}
\begin{tabular}{||c c c ||} 
 \hline
 Metric (50 epochs) & Tiny YOLOv7 & QCFS \\ [0.5ex] 
 \hline\hline
 mAP@.5 & .813 & .836 \\ 
 \hline
 mAP@.5:.95 & .628 & .649 \\
 \hline
 F1 Score & .80 & .82 \\
 \hline
 Total Loss & .041 & .076 \\
 \hline
\end{tabular}
\end{center}

\bibliography{references}  






\end{document}